 \documentclass[pmlr,twocolumn,10pt]{jmlr} % W&CP article

\usepackage[compact]{titlesec}
\titlespacing{\section}{0pt}{2ex}{1ex}
\titlespacing{\subsection}{0pt}{1ex}{0ex}
\titlespacing{\subsubsection}{0pt}{0.5ex}{0ex}

\usepackage{xcolor}
\usepackage{color}
\usepackage{soul}
\usepackage{booktabs}
\usepackage[load-configurations=version-1]{siunitx} % newer version 
\usepackage{chngcntr}
\counterwithin{figure}{section}
\counterwithin{table}{section}

\usepackage{multirow}
\usepackage{makecell}

\theorembodyfont{\upshape}
\theoremheaderfont{\scshape}
\theorempostheader{:}
\theoremsep{\newline}

\jmlrvolume{LEAVE UNSET}
\jmlryear{2021}
\jmlrsubmitted{LEAVE UNSET}
\jmlrpublished{LEAVE UNSET}
\jmlrworkshop{Machine Learning for Health (ML4H) 2021} % W&CP title

\title[Contrastive Learning for Clinical Wearable Data]{Evaluating Contrastive Learning on Wearable Timeseries for Downstream Clinical Outcomes}

\author{%
\Name{Kevalee Shah} \Email{ks877@cam.ac.uk} \\
\Name{Dimitris Spathis} \Email{ds806@cam.ac.uk} \\
\Name{Chi Ian Tang} \Email{cit27@cam.ac.uk} \\
\Name{Cecilia Mascolo} \Email{cm542@cam.ac.uk} \\
\addr{Department of Computer Science and Technology, University of Cambridge}
    \\[-3.2ex]
}

\begin{document}

\maketitle

\begin{abstract}
 Vast quantities of person-generated health data (wearables) are collected but the process of annotating to feed to machine learning models is impractical. This paper discusses ways in which self-supervised approaches that use contrastive losses, such as \textit{SimCLR} and \textit{BYOL}, previously applied to the vision domain, can be applied to high-dimensional health signals for downstream classification tasks of various diseases spanning sleep, heart, and metabolic conditions. To this end, we adapt the data augmentation step and the overall architecture to suit the temporal nature of the data (wearable traces) and evaluate on 5 downstream tasks by comparing other state-of-the-art methods including supervised learning and an adversarial unsupervised representation learning method. We show that SimCLR outperforms the adversarial method and a fully-supervised method in the majority of the downstream evaluation tasks, and that all self-supervised methods outperform the fully-supervised methods. This work  provides a comprehensive benchmark for contrastive methods applied to the wearable time-series domain, showing the promise of task-agnostic representations for downstream clinical outcomes. 
\end{abstract}
\begin{keywords}
Contrastive learning, Self-supervised learning, Time-series data, Mobile health
\end{keywords}

\section{Introduction}
\label{sec:intro} 
    The current bottleneck for supervised learning on data from wearables is the lack of labelled datasets. Features extracted from self-supervised methods can be leveraged in downstream disease classification tasks. Through this we can see how user-generated data can be used to predict user-specific diseases and conditions, which would aid healthcare through early diagnosis.  Therefore, there is a clear motivation to use self-supervised learning to extract features from unlabelled datasets which can be leveraged in downstream classification tasks. Early diagnosis of conditions can lead to a better understanding of the prognosis and can lessen the burden of the healthcare system. Here, we adapt and apply two self-supervised methods,   SimCLR and BYOL, to learn features from a wearable activity dataset, and then use downstream classification tasks of different medical conditions to evaluate the quality of learned representations.

\section{Related Work}
\label{sec:relatedwork}

\paragraph{Self-Supervised Frameworks.}SimCLR and BYOL are both frameworks that first originated for the image domain.   SimCLR is a `simple framework for contrastive self-supervised learning of visual representations' \cite{chen2020simple} that uses data augmentations to create positive and negative samples for the training of the loss function to get high quality learned representations. This is done by using transformations to get different views of the same image, and ensuring that the representations of the positive pairs are attracted and the representations of pairs of views from different images are repelled. In the image domain, the transformations correspond to functions that crop, rotate, resize, colour distort and add noise to the images.
            
`Bootstrap your Own Latent' (BYOL) \cite{Grill2020} presents a new approach to self-supervision that is  simpler and does not require negative samples for the loss function, which has often been the downfall of   SimCLR \cite{arora2019theoretical}. It uses two neural networks working in tandem to generate representations. In the context of image classification, both of these methods have been shown to perform on par with or better than the current state-of-the-art supervised learning methods.

\paragraph{Health Signals.} Recent work has been done in using these self-supervised methods on time-series data, instead of image data. For example,   SimCLR has been used to achieve results that show improvements over supervised and other unsupervised learning methods in the realm of Human Activity Recognition (HAR) \cite{DBLP:journals/corr/abs-2011-11542}, emotion recognition \cite{sarkar2020self} and ECG signals \cite{mohsenvand2020contrastive}. Within the scope of wearable activity data, an adversarial unsupervised representation learning method called \emph{activity2vec} has been used to learn and summarise activity time-series data \cite{aggarwal2019adversarial}. This paper will be referred to as the Adversarial paper henceforth. Their method learns distributed representations for activity signals that span over a time segment in a subject invariant manner. Their work shows that it is possible for representations from self-supervised methods to outperform fully-supervised methods when using linear classifiers for disorder prediction tasks. 
% .  \cite{aggarwal2019adversarial}, that is able to learn and summarise activity time-series data

\section{Data Pre-processing}
    To benchmark all models, the selected dataset is the Hispanic Community Health Study (HCHS) from the National Sleep Research Resource (NSSR) \cite{zhang2018national}.  $1887$ people with Latino origin between the ages of $18$ and $75$ had their activity data measured using a Philip's Actiwatch Spectrum wristwatch\footnote{\url{https://www.usa.philips.com/healthcare/product/HC1046964/actiwatch-spectrum-activity-monitor}} for 7 consecutive days. The time-series for each participant was sampled every $30$ seconds, and metrics such as mean activity count, sleep or awake state and light levels were measured. A visualisation of the data is seen in Figure \ref{fig:hchsdatavisualisation}. Clinical annotations were provided that denote the insomniac, sleep apneic, diabetic, hypertension and metabolic syndrome status of each participant. Disease prevalence statistics are shown in Table \ref{tab:hchs_disease_prevalence}. Pre-processing steps  shown in Figure \ref{fig:datapreprocessingpipeline} were carried out in order to prepare the data for the self-supervised training and downstream classification tasks.   
    
\section{Methodology}
    We adapt SimCLR and BYOL, to suit the time-series nature of the HCHS Dataset. 

\subsection{SimCLR}
    \label{simclr}

        \begin{figure*}[htbp]
         % Caption and label go in the first argument and the figure contents
         % go in the second argument
        \floatconts
          {fig:simclrbyolcombined}
          {\caption{Components of the SimCLR and BYOL frameworks.}}
          {\includegraphics[scale=0.76]{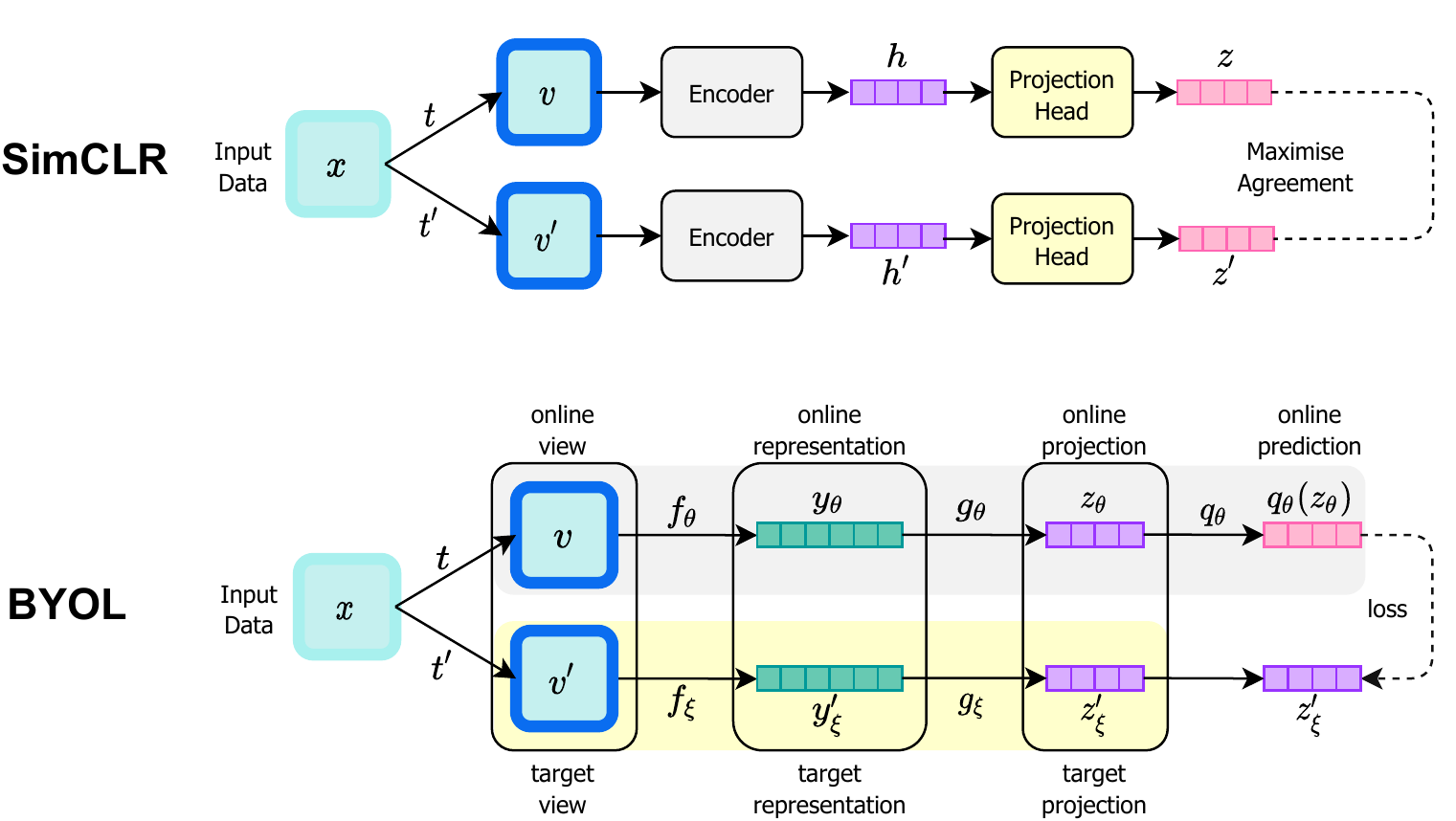}}
    \end{figure*}
        
        The SimCLR framework, originally proposed for image data, is able to efficiently learn useful representations without requiring specialised architectures or a memory bank. The framework is made up of $4$ major components, shown in Figure \ref{fig:simclrbyolcombined}. Specific details of each component can be found in Appendix \ref{sec:simclrappendix}.

        \paragraph{Data Augmentation.}This transforms any given input data into two correlated views of the same piece of data. Transformations provided by \cite{DBLP:journals/corr/abs-2011-11542} are used for time-series data and include: addition of Gaussian noise, scaling, negation, time reversal, column shuffle, time segment shuffle and time warp. Any combination of transformations can be used, and   SimCLR has been shown to be quite sensitive to the order and type of transformations selected.
        
        \paragraph{Neural Network Base Encoder.} The base encoder is responsible for taking the augmented data as input and returning embeddings $\mathbf{h_i}$. In the original   SimCLR paper, a pretrained ResNet50 architecture was utilised as a base encoder \cite{he2015deep}. However, for health signals, no such pretrained models exist and therefore we use an encoder that consists of three 1D convolution layers with $10\%$ dropout. 
        
        \paragraph{Neural Network Projection Head.}A multilayer perceptron head, with three hidden layers of size $258, 128$ and $50$ is used to map the representations, $\mathbf{h_i}$, to get projections $\mathbf{z_i}$. This additional head introduces a non-linear transformation that helps improve the quality of the learned representation. 
        
        \paragraph{Contrastive Loss Function.} NT-Xent loss with a LARS optimiser, learning rate of $0.3 \times \text{BatchSize}/256$ and a cosine decay schedule is used. 
        
\subsection{BYOL}
    Bootstrap Your Own Latent (BYOL) is a self-supervised framework that uses contrastive learning \textit{without} negative samples. We use two copies of an encoder network, called the online and target networks, to obtain representation pairs, and minimize the contrastive loss between them. Figure \ref{fig:simclrbyolcombined} shows the overall architecture of BYOL. Specific details of each step can be found in Appendix \ref{sec:byolappendix}.
    
    \paragraph{Transformations} The augmentation of data is the first step of the BYOL algorithm, and this is done by applying a sequence of transformations to the input data, to achieve two different views. The transformations chosen were noise addition and scaling and signal negation. 
    
    \paragraph{Encoder and Projector} Since no such pretrained network for wearable time-series exists, like ResNet50 for images, we firstly train an autoencoder on the HCHS train dataset, without the labels, and then use the encoder part of the network as the BYOL encoder (denoted by $f$ in Figure \ref{fig:simclrbyolcombined}). As per the BYOL framework, an MLP head is attached to the encoder, which outputs the projection representation.   
    
    \paragraph{Predictor} The predictor is the part of the framework that differentiates the online and target network. It is simply a linear layer that maps from one vector to another vector in the same space ($\mathbb{R}^{p_d}$, where $p_d$ is the projection dimension). The aim of the predictor is to map from the projection of the online network to the projection of the target.

    \paragraph{Training Step} The training algorithm is illustrated in Figure \ref{fig:byoltrainingstep}. Each input data ($x$) is transformed to get two different views ($v$ and $v'$), which are each passed through the online network and the target network. The loss is calculated as the sum of the mean-squared error between the online prediction of $v$ and the target projection of $v'$ and the mean-squared error between the online prediction of $v'$ and the target projection of $v$. Once the loss has been calculated, the gradients are backpropagated through the \emph{online network} only, and the $\theta$ (online) parameters are adjusted. The final part of the training step is to update the $\xi$ (target) parameters, as they are an exponential moving average of the online parameters.

\section{Results}

    \paragraph{Training and Evaluation Protocol.} We use a $80\%, 10\%, 10\%$ split for train, validation and test sets of the labelled HCHS dataset. For the self-supervised training of the contrastive models, SimCLR and BYOL, we use $100\%$ of the train set without using the labels. We report the mean $F_1$-scores with $95\%$ confidence intervals of 10 runs, with the resulting learned representations evaluated on $5$ downstream linear classification tasks, utilising a logistic regression classifier. The classifier is trained using $100\%$ of the representations corresponding to the labelled train set, and then validated and tested with the entirety of the HCHS labelled validation and test sets.

    \begin{table*}[h!]
        \small\addtolength{\tabcolsep}{-5pt}
        % \begin{tabular}{|*{14}{c|}}  % repeats {c|} 18 times
        \centering
        % \begin{adjustwidth}{-0.5in}{-0.5in}
        \scalebox{0.79}{%
        {\def\arraystretch{1.6}\tabcolsep=3pt
        \begin{tabular}{c*{10}{c}}
            \toprule
            \multirow{3}{*}{Method} & \multicolumn{10}{c}{HCHS} \\ %\cline{2-11}
                                    & \multicolumn{2}{c|}{Sleep Apnea} & \multicolumn{2}{c|}{Diabetes} & \multicolumn{2}{c|}{Insomnia} & \multicolumn{2}{c|}{Hypertension} & \multicolumn{2}{c}{Metabolic Syndrome} \\ %\cline{2-11}
                                    & $F_1$-macro & $F_1$-micro & $F_1$-macro & $F_1$-micro & $F_1$-macro & $F_1$-micro & $F_1$-macro & $F_1$-micro & $F_1$-macro & $F_1$-micro \\ \hline
            \makecell{SimCLR} & $\mathbf{50.8 \pm 1}$ & $\mathbf{91.6 \pm 2}$ & $40.0 \pm 2$ & $48.8 \pm 1$ & $32.0 \pm 4$ & $60.0 \pm 4$ & $\mathbf{47.6 \pm 3}$ & $75.1 \pm 1$ &  $\mathbf{52.7 \pm 4}$ & $\mathbf{66.8 \pm 3}$\\ %\hline
            \makecell{BYOL} & $48.0 \pm 1$ & $91.4 \pm 2$ & $23 \pm 2$ & $\mathbf{50 \pm 3}$ & $25 \pm 1$ & $\mathbf{61 \pm 3}$ & $43 \pm 1$ & $\mathbf{76 \pm 2}$ & $42 \pm 1$ & $65 \pm 2$ \\ %\hline
            
            \makecell{Supervised} & $47.7 \pm 0.4$ & $91.2 \pm 1.5$ & $45.2$ & $41.0$ & $50.7$ & $40.1$ & $43.1$ & $75.9 \pm 2.2$ & $39.7 \pm 0.9$ & $65.8 \pm 2.6$ \\ 

            \makecell{day2vec} & $43.6$ & - & $\mathbf{45.8}$ & $42.5$ & $\mathbf{55.7}$ & $41.4$ & $44.1$ & - & - & - \\ 
            
            \bottomrule
        \end{tabular}}}
        % \end{adjustwidth}
        \caption{The best results achieved for each of the 5 downstream tasks with   SimCLR and BYOL, with day2vec and task-specific CNN results shown for comparison. The mean and $95\%$ confidence intervals of 10 runs is reported, apart from the results taken from \textit{day2vec}, which reports just the mean \cite{aggarwal2019adversarial}.}
        \label{tab:hchs_results}
    \end{table*}
    
    \paragraph{Findings.} Table \ref{tab:hchs_results} shows the final results achieved for five downstream classification tasks using SimCLR and BYOL methods. The transformations used for that   SimCLR framework were negation, time segment permutation, time reversal, channel shuffle and random scaling, whereas for    BYOL the transformations used were Gaussian noise addition, scaling and negation. The batch size used for both frameworks was $1024$, with a window size of $512$. In accordance with what was reported in the original   SimCLR paper, larger batch sizes gave better performance than smaller batch sizes. For a baseline supervised learning comparison, we collate results from the Adversarial paper and our own implementation of a task-specific CNN, showing the highest score for each disease in the table, as well as results of \emph{day2vec} which embeds time-series at the level of a day span \cite{aggarwal2019adversarial}. The $F1-$macro and $F1-$micro scores are reported as the evaluation metrics. 
    
    The overall trend we see that   SimCLR and    BYOL out perform the other techniques in the majority of the downstream classification tasks carried out, with the exception of the $F_1$-macro for the classification of diabetes.   SimCLR and    BYOL both outperform the fully supervised method and the \emph{day2vec} method for sleep apnea, and likewise for metabolic syndrome. The classification of metabolic syndrome is a novel downstream evaluation task which has not been studied in any of the papers mentioned in this paper with promising results seen especially with the   SimCLR method for it's classification. With hypertension,   SimCLR can be observed to be performing the best when looking at the macro metric and    BYOL when considering the micro metric. The impact of the evaluation metric used is an area for further work. 
    
    The classification task of both insomnia and diabetes shows the common trend that BYOL performs significantly better than the task specific CNN and \emph{day2vec} in the micro metric, but for the macro metric the \emph{day2vec} method works better. This could be due to the imbalance in classes with those diseases from the dataset, and that the Adversarial paper optimises for the $F_1$-macro metric which provides as explanation for the better result seen there. However this still shows us the advantage self-supervised methods have over fully supervised methods for extracting meaning from time-series data and in particular for health.
    
   \paragraph{Discussion.} This work serves as a good benchmark for the performance of contrastive loss applied to the wearable time-series domain. The clear conclusion to be drawn from these results, and other related work, is that there is definite meaning extracted from the contrastive learning of time-series data, and these representations can be used to boost the performance of linear classification when compared to other fully supervised methods. We observe that   SimCLR performed better than both    BYOL and \emph{day2vec} in majority of the downstream tasks , but it is also clear that different diseases have different characteristics and therefore it can be reasoned that it is expected that different techniques would be more suited to each of the different conditions being evaluated. 
    
    \section{Future Directions}
    We can develop this work further by applying SimCLR and BYOL to other time-series datasets collected through wearables and evaluating the representations on a wider range of downstream classification tasks. Since different techniques result in different representations of the data, for each disease there would be an optimal representation for its classification. Particularly for health care, we can envision a system that monitors the health signals collected from each user's wearables and phone. A set of contrastive techniques would be used to generate a range of representations, and the best representation for each disease being evaluated would be used for the classification. The advantage of such a system is that we use a combination of techniques to get the best possible accuracy, and through the utilisation of these signals we can monitor continuously and ideally identify the onset of any disease early on so that the best treatment can be carried out in time. 

%\section{Citations and Bibliography}
%\label{sec:cite}

\acks{}
This work is partially supported by Nokia Bell Labs through their donation for the Centre of Mobile, Wearable Systems and Augmented Intelligence to the University of Cambridge. CI.T is additionally supported by the Doris Zimmern HKU-Cambridge Hughes Hall Scholarship and from the Higher Education Fund of the Government of Macao SAR, China. D.S is supported by the Embiricos Trust Scholarship of Jesus College Cambridge, and EPSRC through Grant DTP (EP/N509620/1). The authors declare that they have no conflict of interest with respect to the publication of this work.

\bibliography{jmlr-sample}

\newpage
\appendix

\section{SimCLR Details} 
\label{sec:simclrappendix}
\paragraph{Neural Network Base Encoder} The three convolutional layers consist of $32, 64$ and $96$ feature maps with kernels of $24, 16$ and $8$ respectively. The stride length is $1$. \emph{ReLU} is used as the non-linear activation in all the layers, apart from the output. After the final dropout layer, a global max pooling layer is added in order to aggregate the high-level features.

\paragraph{Contrastive Loss Function} In the training algorithm, a batch of $N$ samples is taken, to which transformations are applied to the $N$ positive pairs, and therefore $2N$ total data points are created per batch. The contrastive loss function treats the samples in the batch that come from the same input data as positive pairs, and the rest of the $2(N-1)$ samples in the batch as negative pairs. 

\section{BYOL Details}
\label{sec:byolappendix}

\paragraph{Transformations} The noise addition transformation was done by taking the noise from a Gaussian distribution with $\mu = 0$ and $\sigma = 0.05$. Scaling was applied to the input data by taking scale factors from the normal distribution with $\mu = 1.0$ and $\sigma = 0.1$. 

\paragraph{Encoder and Projector} A simple network for the autoencoder is used, consisting of eight fully connected layers. ReLU and Sigmoid activations are used between each layer. The MLP head is comprised of a fully connected layer that maps the representation vector to a vector of the size $4096$, followed by a 1D BatchNorm layer, ReLU activation, and a final fully connected layer that outputs the projection. 

\paragraph{Training Step} As with the SimCLR implementation, the LARS optimizer with a cosine decay learning rate schedule is used, with a base learning rate that is scaled to the batch size: $\text{LR} = 0.2 \times \text{BatchSize}/256$. The $\theta$ parameters are adjusted using an Adam optimiser in the backpropagation stage. $\beta$ is the target decay rate for updating $\xi$, and it is usually set to $\beta = 0.99$. 

\section{}

    \begin{table*}[h!]
        \centering
        \begin{tabular}{c|c|c}
            \hline
            \textbf{Disease} & \textbf{Classification} & \textbf{Prevalence} \\
            \hline
            \multirow{3}{*}{Diabetes} & Diabetic & 18.1\% \\
            & Pre-Diabetic & 35.0\% \\
            & Non-Diabetic & 46.9\% \\
            \hline
            \multirow{2}{*}{Sleep Apnea} & No & 91.8\% \\
            & Yes & 8.25\% \\
            \hline 
            \multirow{3}{*}{Insomnia} & Moderate to Severe & 17.7\% \\
            & Subthreshold & 22.5\% \\
            & Not Clinically Significant & 59.8\% \\
            \hline 
            \multirow{2}{*}{Hypertension} & No & 74.9\% \\
            & Yes & 25.1\% \\
            \hline 
            \multirow{2}{*}{Metabolic Syndrome} & No & 66.3\% \\
            & Yes & 33.7\% \\
            \hline
        \end{tabular}
        \caption{Disease and Prevalence of the HCHS Dataset}
        \label{tab:hchs_disease_prevalence}
    \end{table*}

\begin{figure}[h]
         % Caption and label go in the first argument and the figure contents
         % go in the second argument
        \floatconts
          {fig:datapreprocessingpipeline}
          {\caption{Pipeline for data pre-processing}}
          {\includegraphics[scale=0.26]{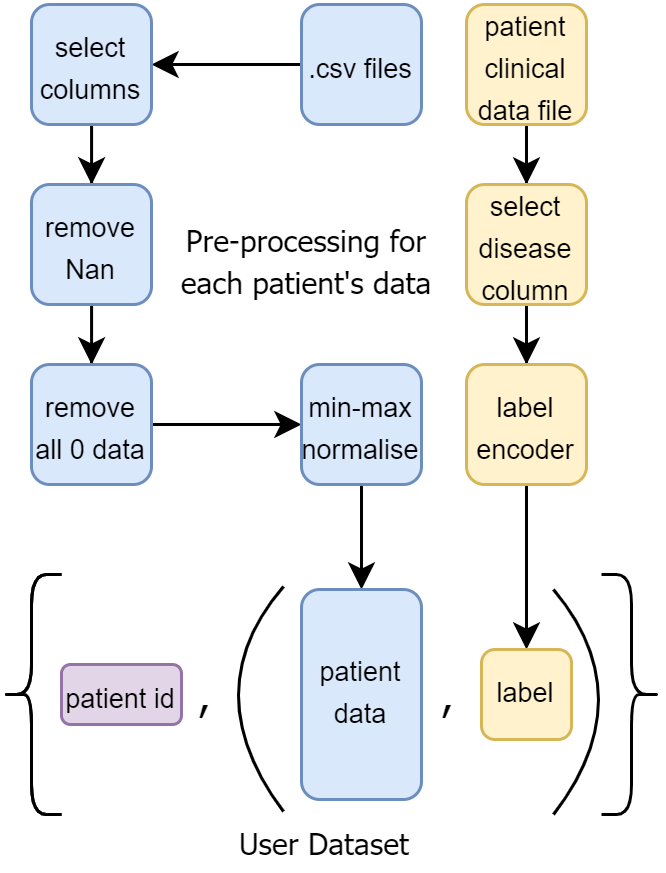}}
    \end{figure}
    
    \begin{figure}[h]
     % Caption and label go in the first argument and the figure contents
     % go in the second argument
    \floatconts
      {fig:byoltrainingstep}
      {\caption{BYOL training step}}
      {\includegraphics[scale=0.3]{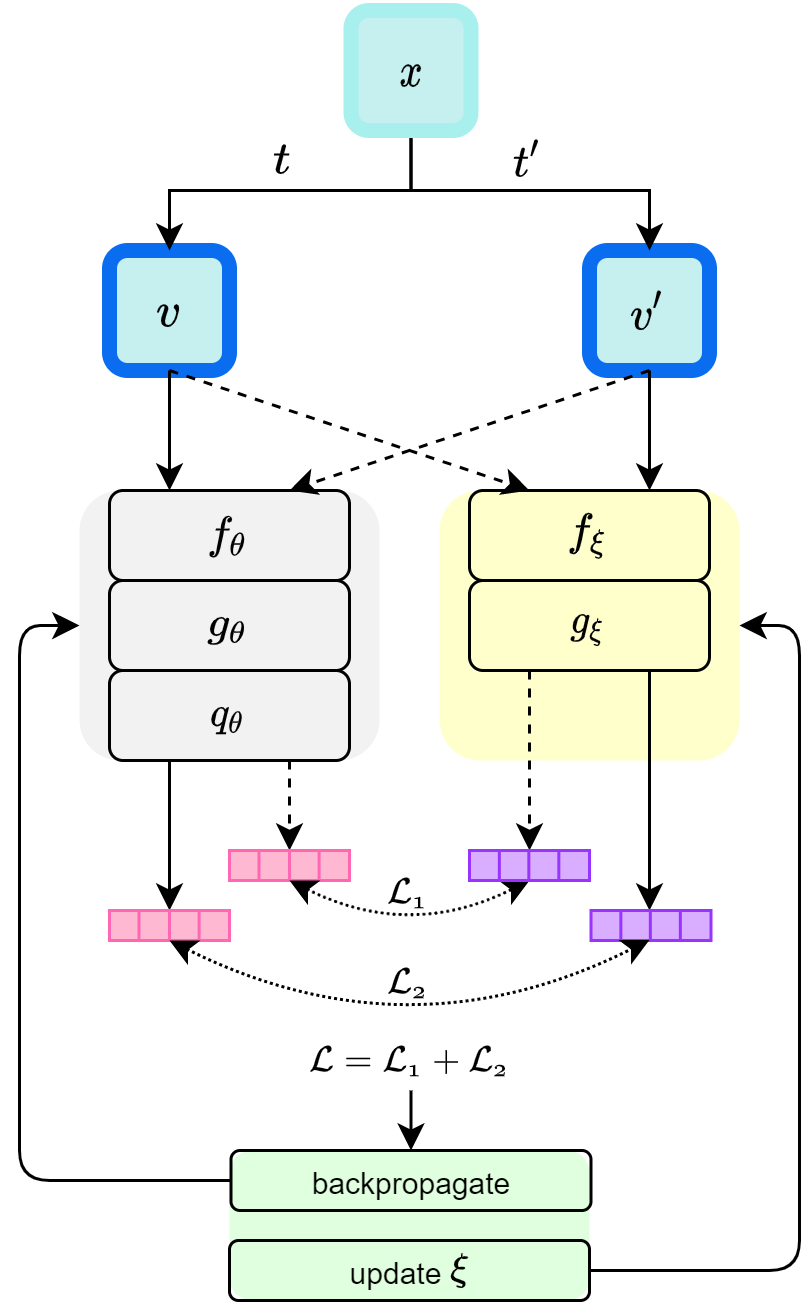}}
    \end{figure}
    
    \begin{center}
    \begin{figure*}[h]
         % Caption and label go in the first argument and the figure contents
         % go in the second argument
         \centering
        \floatconts
          {fig:hchsdatavisualisation}
          {\caption{Plot showing the activity level, light levels and sleep or wake status of Participant 5270581 from the HCHS dataset}}
          {\includegraphics[width=1.1\textwidth]{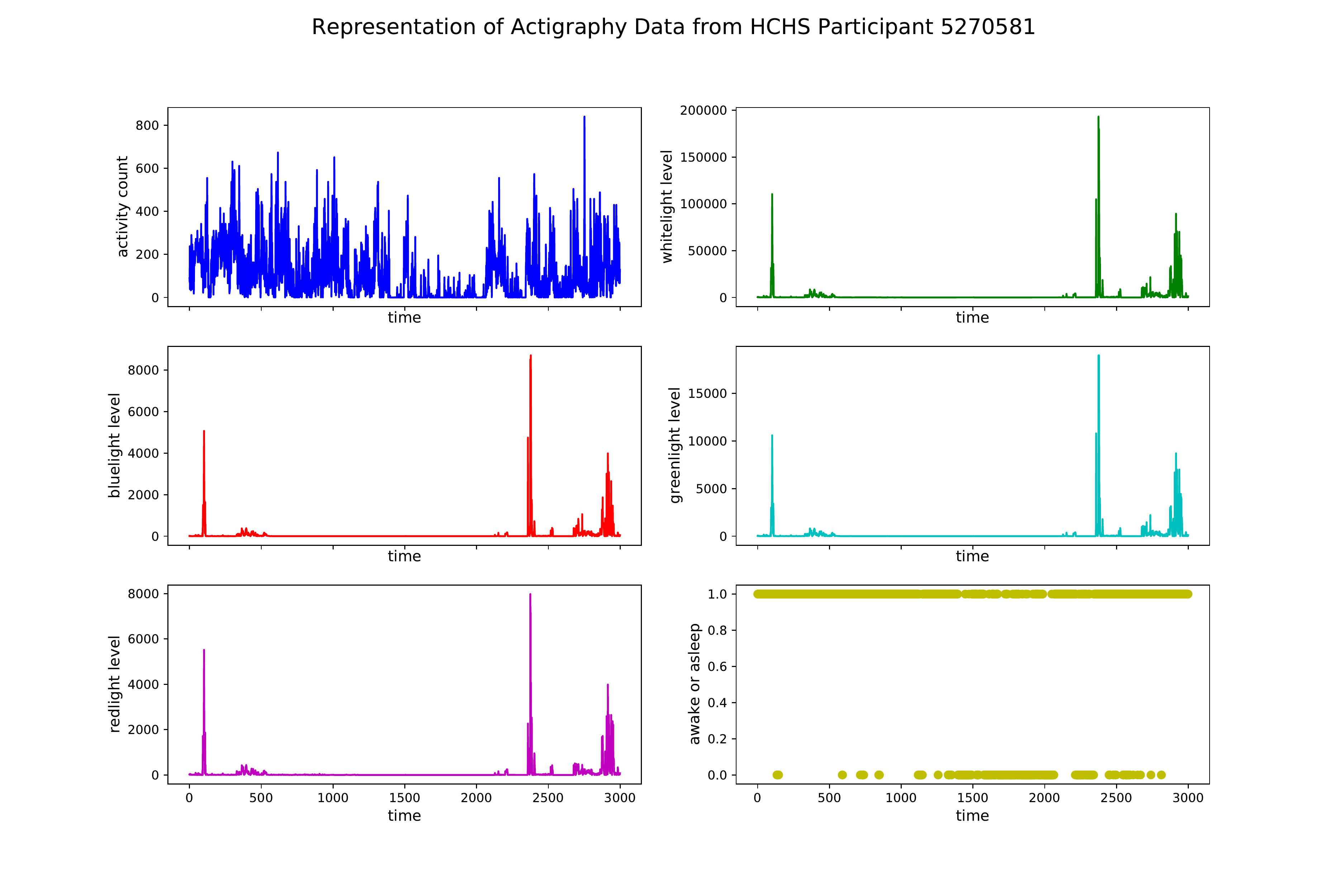}}
    \end{figure*}
    \end{center}

\end{document}